# PAGE: Enriquecimiento de prompts para mejorar la generación de texto

Mauro José Pacchiotti
Universidad Tecnológica Nacional
Centro de I+D de Ing. en Sistemas de Información
Santa Fe, Argentina
mpacchiotti@frsf.utn.edu.ar

Luciana Ballejos
Universidad Tecnológica Nacional
Centro de I+D de Ing. en Sistemas de Información
Santa Fe, Argentina
lballejos@frsf.utn.edu.ar

Mariel Ale
Universidad Tecnológica Nacional
Centro de I+D de Ing. en Sistemas de Información
Santa Fe, Argentina
male@frsf.utn.edu.ar

*Resumen*— En los últimos años, los modelos generativos de lenguaje natural han demostrado un rendimiento sobresaliente en tareas de generación de texto. Sin embargo, cuando se enfrentan a tareas específicas o con requerimientos particulares, pueden presentar rendimientos pobres o necesitar ajustes que requieren grandes cantidades de datos adicionales. Este trabajo propone PAGE (Prompt Augmentation for text Generation Enhancement), un marco de trabajo que permite asistir a estos modelos mediante el uso de módulos auxiliares simples. Estos módulos, modelos simples como clasificadores o extractores, permiten obtener inferencias a partir del texto de entrada. La salida de estos auxiliares se utiliza para construir una entrada enriquecida que permite mejorar la calidad o controlabilidad de la generación. A diferencia de otras propuestas de asistencia a la generación, PAGE no exige el uso de modelos generativos auxiliares, sino que propone una arquitectura más simple, modular y fácil de adaptar a distintas tareas. Este artículo describe la propuesta, sus componentes y arquitectura, y presenta una prueba conceptual en el dominio de ingeniería de requerimientos, donde se utiliza un módulo auxiliar con un clasificador para mejorar la calidad en la generación de requerimientos de software.

*Palabras clave*— Generación de Requerimientos, LLM, Enriquecimiento de Prompts, PAGE

*Abstract*— In recent years, natural language generative models have shown outstanding performance in text generation tasks. However, when facing specific tasks or particular requirements, they may exhibit poor performance or require adjustments that demand large amounts of additional data. This work introduces PAGE (Prompt Augmentation for text Generation Enhancement), a framework designed to assist these models through the use of simple auxiliary modules. These modules—lightweight models such as classifiers or extractors—provide inferences from the input text. The output of these auxiliaries is then used to construct an enriched input that improves the quality and controllability of the generation. Unlike other generation-assistance approaches, PAGE does not require auxiliary generative models; instead, it proposes a simpler, modular architecture that is easy to adapt to different tasks. This paper presents the proposal, its components and architecture, and reports a proof of concept in the domain of requirements engineering, where an auxiliary module with a classifier is used to improve the quality of software requirements generation.

## I. Introducción

La generación de texto se ha convertido en una de las tareas más relevantes dentro del procesamiento de lenguaje natural, gracias a los avances en los grandes modelos lingüísticos (LLM, por sus siglas en inglés) como T5 (Text-to-Text Transfer Transformer) [1], GPT (Generative Pretraining Transformer) [2] o Llama [3] entre otros. Estos modelos, entrenados sobre grandes corpus, son capaces de generar texto con fluidez y coherencia. Sin embargo, cuando se aplican en tareas específicas donde se requiere que la salida cumpla con ciertas condiciones, limitaciones o estilos particulares, los resultados no siempre son favorables.

La problemática descripta -en algunos casos- puede solucionarse con el rentrenamiento o ajuste de parte o la totalidad del modelo. Aunque esta decisión de utilizar el entrenamiento de los modelos para mejorar el desempeño en una tarea tiene dos grandes implicancias. Por un lado, hace falta reunir y conformar conjuntos de datos con la cantidad y calidad suficiente de muestras, y por otro, se requiere del poder de cómputo necesario para la tarea de entrenamiento. Estas necesidades implican la disponibilidad de recursos que a veces no se disponen y, por lo tanto, no se hace posible lograr el objetivo propuesto.

Frente a las dificultades y necesidades descriptas, este trabajo propone PAGE, una arquitectura que incorpora módulos auxiliares que permiten obtener inferencias del texto de entrada. Estos módulos auxiliares pueden ser clasificadores, analizadores o extractores de características, entre otras opciones. Al ejecutarse antes del modelo generativo, los modelos auxiliares aportan metadatos o información estructurada que se incorpora a la entrada del generador. La arquitectura es modular y puede adaptarse según la tarea a resolver, permitiendo combinar distintos tipos de módulos auxiliares según las necesidades del dominio y la tarea a realizar.

El aporte de los módulos auxiliares pretende mejorar la salida del modelo generador y consumir menos recursos, tanto en el uso para generación, como también para el entrenamiento, ya que pueden usarse auxiliares simples que no requieren de gran poder de cómputo o grandes conjuntos de datos de entrenamiento.

El resto del trabajo está organizado de la siguiente manera: la Sección II presenta el marco teórico y trabajos relacionados,

la Sección III describe la herramienta PAGE, la Sección IV desarrolla una prueba conceptual centrada en la generación de requerimientos estructurados con sintaxis EARS [4]. Finalmente, en la Sección V se informan los resultados, y en la Sección VI se exponen las conclusiones y se proponen posibles líneas de trabajos futuros.

## II. Marco Teórico y Trabajos Relacionados

En tiempos recientes surgieron algunas propuestas para mejorar la entrada a un modelo generativo en busca de una mejor salida. Si bien varias propuestas utilizan modelos para asistir al generador, son variadas las formas de aplicación posibles. En esta línea, Du et al. [5] propone mejorar LLMs en tareas de inferencia textual combinando prompts explícitos con conocimiento semántico extraído de bases externas. Utilizan atributos inferidos como insumos auxiliares, lo que mejora la precisión y coherencia de las respuestas. Por otro lado, He et al. [6] propone asistir la generación de GPT-2 mediante resúmenes humanos codificados con BERT, guiando así la generación hacia mayor coherencia temática. Se evalúan distintas arquitecturas híbridas, con mejoras moderadas. En otro trabajo Zeldes et al. [7] introducen Auxiliary Tuning, una técnica que adapta LLMs preentrenados a nuevas tareas, agregando un modelo auxiliar que ajusta la distribución de salida. La combinación se realiza a nivel de logits, sin modificar los pesos originales.

Más recientemente, Zhang et al. [8] proponen IAG (Induction-Augmented Generation), donde se utiliza un modelo generativo auxiliar para inducir conocimiento a partir del contexto, que luego se incorpora como entrada a otro modelo generativo. Esta inducción ha demostrado buenos resultados en tareas de razonamiento y QA. En otra propuesta, Liao et al. [9] proponen Awakening Augmented Generation, una técnica que activa el conocimiento latente en LLMs mediante tareas auxiliares previas, mejorando respuestas en QA. Su enfoque demuestra que pequeñas intervenciones bien diseñadas pueden guiar la generación sin alterar los parámetros base.

A diferencia de estos trabajos, PAGE propone utilizar módulos auxiliares simples y construir con sus salidas una entrada enriquecida que el modelo generativo pueda utilizar para mejorar sus respuestas. Esto permite una mayor interpretabilidad, modularidad y adaptabilidad, haciendo posible su implementación en escenarios con recursos limitados.

### A. EARS

La propuesta de EARS [4] se basa en la identificación de patrones recurrentes en los requerimientos. A partir de un análisis empírico de especificaciones reales, los autores establecieron un conjunto reducido de plantillas sintácticas que cubren la mayoría de los casos prácticos. Estas plantillas permiten expresar diferentes categorías de requerimientos: Ubiquitous, Event-driven, State-driven, Optional y Unwanted, utilizando una sintaxis clara, que guía al analista en la redacción de cada expresión. De este modo, se logra un lenguaje controlado que reduce la ambigüedad, sin exigir conocimientos técnicos avanzados en lenguajes formales.

Uno de los beneficios principales de EARS es que facilita la comunicación entre stakeholders. Al proporcionar una estructura reconocible y repetible, se reduce la presencia de ambigüedad y se mejora la trazabilidad de los requerimientos a lo largo del ciclo de vida del software. Asimismo, la simplicidad de la técnica permite que usuarios no especializados participen en la redacción y revisión de las especificaciones.

### B. Grandes Modelos Lingüísticos

A partir del trabajo *Attention is all you need* [10] surge el Transformer, un modelo de aprendizaje profundo cuya arquitectura se organiza en dos estructuras principales: un codificador y un decodificador, ambos basados en el mecanismo de atención multicabeza. A diferencia de los modelos recurrentes que procesan el texto palabra por palabra en orden secuencial, el Transformer puede considerar todas las palabras de una oración al mismo tiempo. Gracias a este mecanismo, los modelos lingüísticos pueden resaltar las partes más relevantes de una entrada y comprender mejor tanto su significado como su contexto. Esto hace posible capturar relaciones de largo alcance entre términos, lo cual se traduce en mejoras sustanciales en múltiples tareas de procesamiento del lenguaje natural.

A partir del Transformer se desarrollaron distintos modelos que adoptaron y expandieron esta arquitectura, como BERT (Bidirectional Encoder Representations from Transformers) [11], T5 [1], GPT [2] y Llama [3] entre otros. Estos modelos se entrenaron con volúmenes cada vez mayores de datos, incluyendo texto y código, y se destacaron por su capacidad para generar texto de alta calidad y adaptarse a un amplio rango de tareas en PLN. La evolución continuó con la aparición de ChatGPT [12] en 2022. Esta aplicación de OpenAI llevó el uso de la IA generativa a un público amplio a través de una interfaz de chat, basada en la familia de modelos GPT, capaz de generar respuestas coherentes y contextuales en lenguaje natural.

### C. Técnicas de Prompting

De esta interacción con LLMs surge la posibilidad de comunicarse con un modelo mediante expresiones en lenguaje natural. En este contexto, la entrada que el usuario proporciona recibe el nombre de prompt, entendido como una instrucción o conjunto de palabras que orientan la generación de la respuesta. La manera en que se formula este prompt resulta fundamental, ya que condiciona directamente la salida producida por el modelo. Por esto, distintas guías de buenas prácticas señalan qué elementos conviene considerar al redactarlo con el fin de obtener resultados más cercanos a lo esperado (ver Tabla I).

TABLA I. ELEMENTOS RECOMENDADOS EN UN PROMPT. [13]

| Elemento | Descripción |
| --- | --- |
| Instrucción | Tarea específica que se desea. |
| Contexto | Información adicional que puede orientar al modelo y completar la respuesta. |
| Entrada | La entrada sobre la que se desea la acción. |
| Salida | Formato que se desea para la respuesta del modelo. |

Es importante destacar que estos elementos no son estrictamente obligatorios, sino que su inclusión depende de la necesidad en cada caso. Asimismo, el prompt puede enriquecerse incorporando ejemplos de la tarea deseada, lo que permite al modelo imitar de manera más precisa el comportamiento esperado. De acuerdo con la cantidad de ejemplos aportados, se reconocen tres enfoques principales: zero-shot (sin ejemplos), one-shot (con un ejemplo) y few-shot (con varios ejemplos). Entre ellos, el enfoque few-shot suele ser el más potente, ya que mejora la capacidad de generalización del modelo y produce salidas de mayor calidad.

### D. ROUGE

La evaluación automática de sistemas de generación de texto requiere métricas que permitan medir la calidad de una salida en comparación con referencias humanas. En este ámbito, ROUGE (Recall-Oriented Understudy for Gisting Evaluation), propuesta por Lin [14], se consolidó como una de las técnicas más utilizadas en el análisis de resúmenes automáticos y se extiende a la evaluación de modelos generativos. La métrica se basa en la superposición de unidades de texto (n-gramas, secuencias o subsecuencias) entre el texto generado y uno o más textos de referencia, midiendo así la similitud entre ellos.

Entre las variantes más empleadas se encuentran ROUGE-1, ROUGE-2 y ROUGE-L. ROUGE-1 evalúa la coincidencia de unigramas (palabras individuales) entre el resultado y la referencia, proporcionando una medida básica de cobertura del contenido. ROUGE-2, por su parte, se centra en la coincidencia de bigramas, lo que introduce un nivel mayor de sensibilidad al orden y la fluidez de las palabras, capturando relaciones locales entre términos. Finalmente, ROUGE-L se basa en la subsecuencia más larga de palabras en común entre las cadenas comparadas (LCS En inglés: Longest Common Subsequence), lo que permite valorar la preservación de la estructura y el orden global de la información [14].

Un aspecto importante de ROUGE es que puede calcularse en términos de recall (1), precisión (2) y F1-score (3), aunque en el ámbito de generación de texto se utiliza más frecuentemente el recall, al priorizar la recuperación del contenido presente en el texto de referencia.

$$Recall_{ROUGE} = \frac{n-gramas\ coincidentes}{n-gramas\ en\ la\ referencia} \quad (1)$$

$$Precision_{ROUGE} = \frac{n-gramas\ coincidentes}{n-gramas\ en\ la\ generación} \quad (2)$$

$$F1Score_{ROUGE} = \frac{2.Precicision.Recall}{Precision+Recall} \quad (3)$$

## III. HERRAMIENTA PROPUESTA

La propuesta PAGE parte de la idea de que los modelos generativos pueden beneficiarse al recibir como entrada no sólo el texto original, sino también información extra estructurada inferida del mismo texto. Estas inferencias pueden obtenerse mediante módulos auxiliares específicos, modelos o algoritmos, diseñados según la tarea de generación a resolver. Estos módulos auxiliares permiten, a partir de sus inferencias mejorar la salida del modelo generador. Si se analiza desde la perspectiva de los recursos, se traduce en una mejora en la tarea de generación sin entrenar o ajustar un LLM, sino pequeños modelos auxiliares. Se trata de una alternativa que pretende mejorar la generación de un modelo reduciendo la necesidad de grandes conjuntos de datos y capacidades de cómputo para realizar ajustes o entrenamientos.

PAGE se estructura como una arquitectura modular compuesta por tres componentes principales: el conjunto de módulos auxiliares, el compositor de contexto, y el generador principal. Cada componente puede adaptarse según la aplicación concreta, permitiendo utilizar módulos simples y muy interpretables como clasificadores, etiquetadores, extractores o analizadores de sentimientos, entre otras opciones. También pueden ser útiles funciones que aporten de acuerdo con la expresión original, información de alguna fuente externa.

### A. Módulo Auxiliar

El módulo auxiliar es el componente responsable de realizar una o más inferencias sobre el texto de entrada, con el fin de obtener información estructurada que luego será utilizada para asistir a la generación. A diferencia del modelo generativo principal, estos modelos son generalmente más simples, entrenados para tareas específicas y diseñados para ser fácilmente interpretables. Además, al ser modelos más sencillos requieren un menor esfuerzo de entrenamiento, tanto desde el punto de vista del poder de cómputo como de los datos. El tipo de modelo auxiliar a emplear depende directamente del dominio y los objetivos de generación. A continuación, se describen algunos tipos comunes:

- *Clasificador:* Un modelo que clasifica con alguna etiqueta de interés para la tarea el texto de entrada. Existe una gran variedad de modelos y pipelines que pueden realizar esta tarea, utilizando desde simples modelos estadísticos hasta redes neuronales profundas o grandes modelos lingüísticos.

- *Extractor de entidades o partes del discurso:* Este tipo de modelo identifica y clasifica entidades, acciones o porciones relevantes dentro del texto. En estos casos la salida podría tratarse de una estructura con los tokens extraídos.

- *Analizador de sentimiento o intención:* Se trata de un tipo de clasificador que permite detectar el tono, la urgencia o la finalidad de una necesidad, así como también el sentimiento que expresa el autor del texto. Existen varias propuestas para las etiquetas de salida en estos casos.

En todos los casos, la salida del módulo auxiliar debe ser expresada de forma explícita y legible, generalmente como texto estructurado, de modo que pueda ser utilizada sin preprocesamientos por el Compositor de Prompts. Esta estrategia facilita la trazabilidad del sistema manteniendo interfaces claras, lo que resulta importante para las actividades de gestión y control del proceso.

### B. Compositor de Prompts

Este artefacto toma la salida de los módulos auxiliares, que pueden ser uno o más, y construye un bloque de entrada mejorado que se combina con el texto original. Esta composición puede seguir plantillas configurables o estructuras semiformales según el objetivo de la instrucción y las estructuras devueltas por los módulos auxiliares. Finalmente, la salida es el texto que se utiliza como entrada en el modelo generativo.

### C. Modelo Generativo

El componente generativo puede ser cualquier LLM capaz de producir texto a partir de la entrada de un prompt. Si bien puede efectuarse un ajuste fino o reentrenamiento del modelo generativo, este no siempre resulta necesario; se espera que el enriquecimiento contextual, correctamente estructurado, sea suficiente para dirigir la generación hacia una salida con la calidad deseada.

Dependiendo de la aplicación pueden utilizarse distintas técnicas de prompting para utilizar el modelo generativo, dentro de las que se destacan las que tienen que ver con los ejemplos que se proveen al modelo, One-shot y Few-shots.

### D. Proceso

El flujo de trabajo en PAGE se resume en los siguientes pasos:

1) Un usuario o sistema proporciona un texto base.

2) Los módulos auxiliares procesan esta entrada y generan información estructurada a partir de sus inferencias.

3) El compositor integra la información estructurada con el texto original y construye un prompt enriquecido mediante una plantilla.

4) El modelo generativo produce la salida final a partir de esta entrada enriquecida.

El flujo del proceso (Fig. 1) permite realizar pruebas de los componentes por separado para analizar el impacto de cada auxiliar, así como adaptar el marco a distintos dominios sin modificar el modelo generativo. Además, la naturaleza textual de los componentes auxiliares facilita la depuración, interpretación y validación de los pasos intermedios.

### IV. Prueba Conceptual

Para probar la propuesta se diseñó una implementación del marco con el objetivo de mejorar expresiones de requerimientos de software. Son diversas las propuestas sobre la estructura sintáctica que se debe utilizar para expresar requerimientos, con el objetivo de reducir la ambigüedad y otras deficiencias. El enfoque EARS [4] ofrece clasificar los requerimientos de software en cinco categorías: Event-driven, Ubiquitous, State-driven, Unwanted behavior y Optional; para posteriormente emplear una plantilla particular para cada tipo, lo que permite guiar y restringir la generación de las expresiones.

### A. Conjunto de Datos

Las pruebas se realizan sobre un Dataset que resulta de la recopilación de textos de requerimientos desde diversas fuentes, los Datasets *PURE* [15] y *Software Functional Requirements* [16], así como también requerimientos obtenidos desde diversos documentos de especificación de dominio público. Cabe resaltar que se buscó diversidad de dominios y balance con respecto a las categorías de la propuesta EARS [4] (Figura 2).

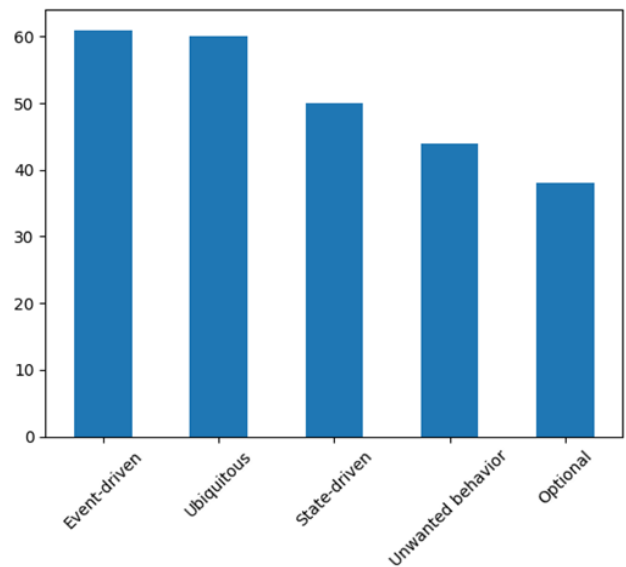

Fig. 2. Balance del conjunto de datos.

La estructura del dataset está compuesta por tres columnas: a) la expresión de requerimiento sin una estructura sintáctica definida, b) la etiqueta correspondiente con la categoría de EARS y c) la expresión con sintaxis EARS elaborada manualmente, conforme a las indicaciones del enfoque. El conjunto de datos empleado consta de 253 instancias, un tamaño reducido que resulta apropiado para evaluar en qué medida la propuesta permite disminuir el esfuerzo asociado a la preparación de conjuntos de datos para entrenamiento.

### B. Componentes

En esta implementación de PAGE, con el fin de generar expresiones de requerimientos según la propuesta EARS, se definen los siguientes componentes:

*Modulo auxiliar:* Para esta implementación se utiliza un solo módulo auxiliar que contiene un clasificador. Este modelo simple, dada una expresión textual, devuelve la etiqueta correspondiente a la categoría EARS. Luego, la etiqueta es utilizada para que el módulo devuelva ejemplos correspondientes con esa categoría que puedan ser anexados como información de contexto al prompt.

Para disponer de un modelo clasificador que pueda entrenarse con pocas muestras, se realizó un entrenamiento con búsqueda de grilla para ajuste de hiperparámetros de un modelo Random Forest [17]. Este tipo de modelos es una de las técnicas de aprendizaje supervisado más utilizadas frente a conjuntos de datos reducidos. Al basarse en un ensamble de árboles de decisión entrenados sobre subconjuntos de datos y características, logra disminuir el riesgo de sobreajuste que suelen presentar los modelos individuales. Esta propiedad lo convierte en una alternativa confiable cuando se dispone de un número limitado de muestras, ya que aprovecha la variabilidad introducida por el muestreo y mantiene un equilibrio entre sesgo y varianza [17]. Además, cumple con una de las motivaciones de la propuesta, por tratarse de un modelo que requiere de muy poco poder de cómputo para su entrenamiento.

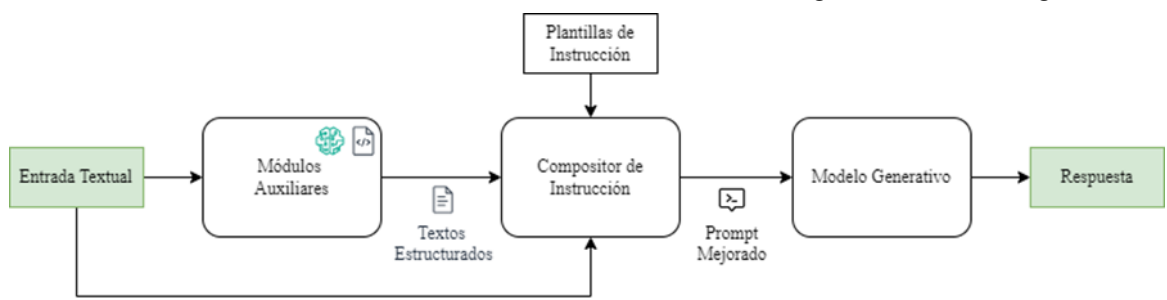

Fig. 1. Proceso de PAGE

La configuración de hiperparámetros que obtuvo el mejor desempeño para el modelo Random Forest correspondió a una profundidad máxima de 10, un mínimo de 5 muestras por división y 100 estimadores. Para el entrenamiento, el conjunto de datos se particionó reservando un 20% para pruebas, complementado con un esquema de validación cruzada de cinco particiones. Con esta configuración, el modelo alcanzó un accuracy del 82.35% sobre el conjunto de test. La Figura 3 muestra las medidas de performance obtenidas y la Figura 4 la matriz de confusión.

```
                  precision    recall  f1-score   support

   Event-driven       0.83      0.83      0.83        12
       Optional       0.88      0.88      0.88         8
   State-driven       1.00      0.80      0.89        10
     Ubiquitous       0.69      0.92      0.79        12
Unwanted behavior     0.86      0.67      0.75         9

       accuracy                           0.82        51
      macro avg       0.85      0.82      0.83        51
   weighted avg       0.84      0.82      0.82        51
```

Fig. 3. Resultados del modelo de clasificación con el conjunto de Test.

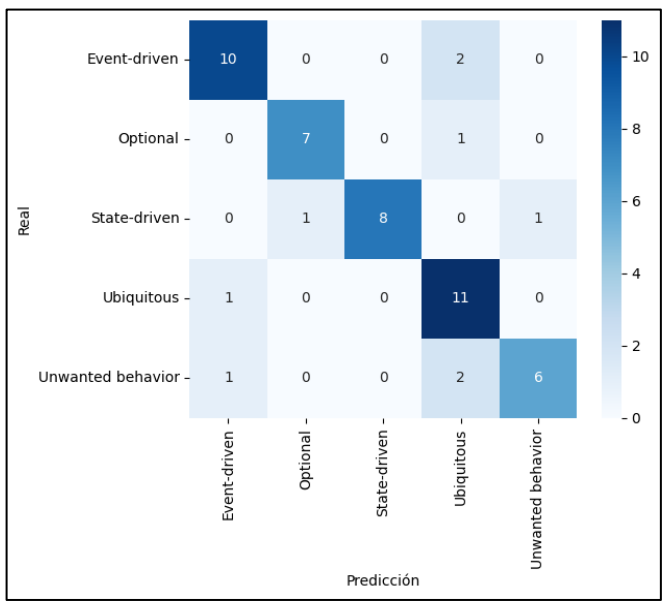

Fig. 4. Matriz de confusión de los resultados con el conjunto de Test.

Esta etiqueta obtenida por el clasificador se utiliza como parámetro para que el módulo devuelva dos ejemplos de requerimientos escritos según la sintaxis EARS para esa categoría. La Tabla II muestra los ejemplos utilizados en las pruebas, de acuerdo con la categoría EARS.

*Compositor de contexto:* El propósito de este componente es generar el prompt mejorado que será ingresado al modelo generativo. Para este ejemplo se utiliza una plantilla (Figura 5) que permite anexar los ejemplos devueltos por el módulo auxiliar en la etiqueta {examples_text}.

TABLA II. EJEMPLOS SEGÚN LA CATEGORÍA EARS.

| Categoría | Ejemplos |
|---|---|
| Ubiquitous | **requirement:** "The system shall log all transactions." **ears:** "The system shall always log all transactions." **requirement:** "The application shall keep the session active during user activity.", **ears:** "The system shall always keep the session active during user activity." |
| Event-driven | **requirement:** "The system shall notify the admin when the server restarts.", **ears:** "When the server restarts, the system shall notify the admin." **requirement:** "The application shall send a receipt when a purchase is completed.", **ears:** "When a purchase is completed, the application shall send a receipt." |
| State-driven | **requirement:** "The system shall block new logins while maintenance mode is active." **ears:** "While maintenance mode is active, the system shall block new logins." **requirement:** "The application shall allow offline access while the device has no internet connection." **ears:** "While the device has no internet connection, the application shall allow offline access." |
| Unwanted behavior | **requirement:** "The system shall display a warning if unauthorized access is detected.", **ears:** "If unauthorized access is detected, the system shall display a warning.". **requirement:** "The application shall stop the upload if the file exceeds the maximum size.", **ears:** "If the file exceeds the maximum size, the application shall stop the upload." |
| Optional | **requirement:** "The system shall enable voice control where the device supports it.", **ears:** "Where the device supports it, the system shall enable voice control." **requirement:** "The application shall provide dark mode where the user has selected the option.", **ears:** "Where the user has selected the option, the application shall provide dark mode." |

*Modelo generativo:* En esta implementación se utiliza un modelo generativo con licencia de uso público, Llama 3.1[1] en su versión con 8 billones de parámetros entrenables. Para implementarlo se despliega sobre la herramienta Ollama[2] que permite ejecutarlo de manera local y consumirlo como un servicio desde un entorno Jupyter Notebook[3] con el lenguaje de programación Python[4].

*Experimentos:* Se realizaron tres pruebas sobre el dataset completo, 253 filas, con el objetivo de recolectar métricas que permitan validar la utilidad de la propuesta. La primera consistió en utilizar únicamente el modelo generativo junto con la expresión textual original y un prompt sin optimizaciones, lo que permitió establecer una línea base de desempeño. En la segunda, se incorporó el Compositor de Prompts y una versión ideal del Módulo Auxiliar, que disponía de la etiqueta correcta para cada expresión proveniente del dataset. Este diseño permitió obtener simultáneamente el piso de desempeño (sin enriquecimiento del prompt) y el techo alcanzable (con enriquecimiento

---

[1] https://huggingface.co/meta-llama/Llama-3.1-8B
[2] https://ollama.com/
[3] https://jupyter.org/
[4] https://www.python.org/

```
You are an assistant that rewrites requirements using the EARS syntax.
Rewrite the following requirement using the EARS syntax.
Use the examples below as a guide.
{examples_text}
------------------------------------------------------------------------
Respond ONLY with the rewritten requirement. Do not add explanations, comments, or any extra text.Requirement:

{natural}

EARS Requirement:
```

Fig. 5. Plantilla para generación del prompt.

derivado de las etiquetas correctas provistas por el conjunto de datos). Finalmente, la tercera prueba implementó el proceso PAGE completo, utilizando el Módulo Auxiliar descripto en la Sección 4.2.

## V. RESULTADOS

Para evaluar la propuesta se compararon las expresiones generadas en las tres pruebas realizadas. Como métricas para comparar el desempeño se utilizaron ROUGE 1, ROUGE 2 y ROUGE L calculando para cada una el recall, la precisión y el F1-Score, comparando cada expresión con la correcta, provista por el dataset. La Tabla III muestra los resultados obtenidos en cada prueba.

TABLA III. RESULTADOS PARA EL MODELO SIN MÓDULOS AUXILIARES (ZERO-SHOT) CON MÓDULO AUXILIAR BASADO EN LA ETIQUETA DEL DATASET (DATASET-SAMPLES) Y PARA LA PROPUESTA PAGE (PAGE)

|  | Métrica | Precisión | Recall | F1-Score |
|---|---|---|---|---|
| Zero-Shot | ROUGE1 | 0,509 | 0,489 | 0,485 |
| Zero-Shot | ROUGE2 | 0,206 | 0,204 | 0,199 |
| Zero-Shot | ROUGEL | 0,413 | 0,395 | 0,392 |
| Dataset-samples | ROUGE1 | 0,852 | 0,815 | 0,827 |
| Dataset-samples | ROUGE2 | 0,653 | 0,630 | 0,636 |
| Dataset-samples | ROUGEL | 0,803 | 0,770 | 0,781 |
| PAGE | ROUGE1 | 0,849 | 0,809 | 0,822 |
| PAGE | ROUGE2 | 0,648 | 0,622 | 0,630 |
| PAGE | ROUGEL | 0,796 | 0,761 | 0,772 |

En los resultados se destacan los valores altos logrados con la métrica ROUGE 1, lo que sugiere que los modelos están capturando bien los términos individuales, aunque también logran reproducir combinaciones locales de palabras (bigramas) y estructuras más largas (ROUGE-L). La Figura 6 muestra gráficamente las diferencias en los rendimientos alcanzados en las pruebas.

También se realiza un análisis para medir la mejora de la implementación de PAGE con respecto a la línea base lograda con el modelo sin módulos auxiliares y el óptimo utilizando un auxiliar que entrega la etiqueta correcta desde el dataset, la Tabla IV muestra los porcentajes de mejora sobre el Recall de las métricas ROUGE.

TABLA IV. PORCENTAJES DE MEJORA SOBRE LA LÍNEA BASE

|  | Data-Model | PAGE |
|---|---|---|
| Rouge 1 | 66,72% | 65,41% |
| Rouge 2 | 209,54% | 205,62% |
| Rouge L | 95,21% | 92,79% |

Como puede observarse, la implementación de PAGE sólo queda entre 2 y 4 puntos porcentuales por debajo del ideal, lo que demuestra el buen desempeño de la propuesta.

Finalmente, se realiza un análisis cualitativo de las salidas obtenidas en las tres pruebas. Éste permite comprobar la alta semejanza entre las salidas de la implementación ideal, con etiquetas del Dataset y la implementación de PAGE. La Tabla V muestra ejemplos de las distintas expresiones generadas, donde puede observarse que en el caso base -donde se usa el prompt sin ejemplos-, el modelo devuelve una expresión totalmente distinta a la esperada.

Esto se debe a que, sin ejemplos de referencia, el modelo tiende a realizar la generación de la manera más general,

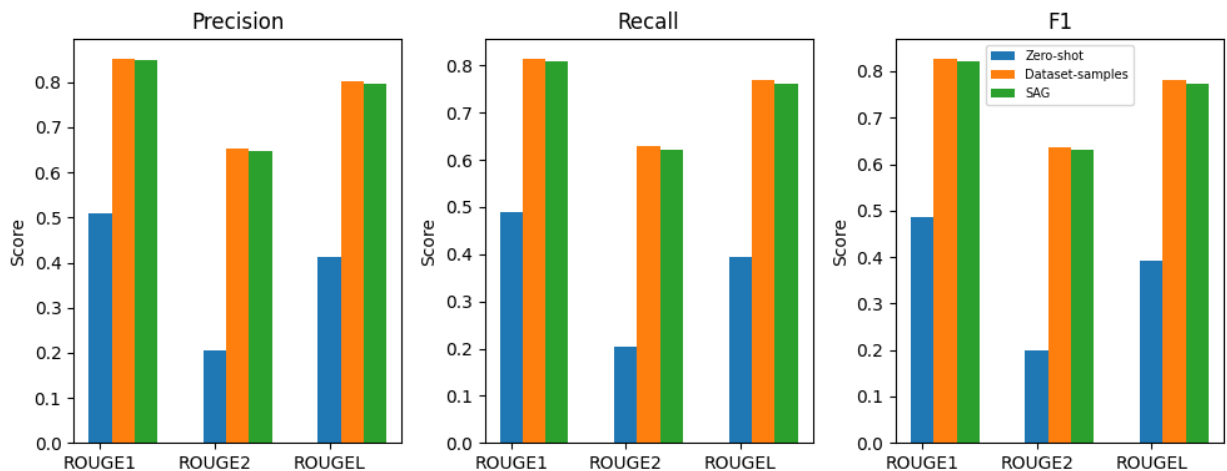

Fig. 6. Rendimientos obtenidos en las tres pruebas realizadas.

buscando una estructura coherente según su conocimiento previo adquirido durante el preentrenamiento, en lugar de seguir el patrón sintáctico esperado. En ausencia de un marco de instrucción claro, el modelo puede optar por representaciones estructuradas del requerimiento, como la identificación de actores, eventos o roles, porque asocia ese tipo de salida con tareas de la ingeniería de requerimientos. Por otro lado, al incorporar ejemplos dentro del prompt, el modelo dispone de un contexto semántico y sintáctico más definido, lo que reduce la ambigüedad sobre el formato de salida esperado.

De este análisis puede concluirse que el módulo auxiliar tiene gran influencia sobre la generación, así, cuando la clasificación falla, conduce al modelo generativo a una estructura sintáctica equivocada para esa clase de requerimiento. Esto demuestra que, en las implementaciones del marco, es importante dedicar esfuerzos a mejorar las respuestas de los módulos auxiliares para lograr aportes correctos al modelo generativo.

TABLA V. Ejemplos de las generaciones obtenidas en las distintas pruebas.

| Ejemplo | Origen | PAGE |
|---|---|---|
| 0 | Salida esperada | The system shall allow a customer to place an order online. |
| 0 | Zero-shot | Event: Customer places an order<br>Actor: The customer<br>Role: Online shopper<br>Scope: Order placement |
| 0 | Dataset-label | The system shall always allow a customer to place an order online. |
| 0 | PAGE | The system shall always allow a customer to place an order online. |
| 9 | Salida esperada | The system shall provide reports of actions on cases organized by case, by user, and in chronological sequence. |
| 9 | Zero-shot | The system shall generate reports that present actions on cases in the following formats:<br>* E: Each report shall display data organized by:<br>    + Case<br>    + User<br>* A: And/or<br>* R: Reports shall be generated in a chronological sequence. |
| 9 | Dataset-label | The system shall always provide reports for actions on cases, organized by case, by user, and in chronological sequence. |
| 9 | PAGE | The system shall always provide reports for actions on cases, organized by case, by user, and in chronological sequence. |
| 91 | Salida esperada | When a driver completes a ride, the system shall allow the driver to leave a review. |
| 91 | Zero-shot | The Driver shall be enabled to submit a review after successfully completing a ride. |
| 91 | Dataset-label | When a ride is completed, the Application shall enable the driver to leave a review. |
| 91 | PAGE | When a ride is completed, the Application shall enable the driver to leave a review. |

## VI. Conclusiones y Trabajos Futuros

PAGE propone un enfoque sencillo y eficaz para mejorar el rendimiento, la controlabilidad y la estructura de las salidas generadas por LLMs, aprovechando inferencias y aportes simples y adaptables al contexto de uso. Los resultados obtenidos en las pruebas iniciales son prometedores y abren la posibilidad de replicar los experimentos con otros datasets y en diferentes dominios. Se concluye que la solución planteada logra medidas alentadoras para las métricas evaluadas, lo que sugiere la conveniencia de ampliar la validación hacia conjuntos de datos con más requerimientos y en diferentes dominios.

El principal aporte de esta propuesta radica en demostrar que la generación de texto puede manipularse de manera efectiva mediante herramientas sencillas, apoyadas en módulos auxiliares simples y altamente interpretables. Estos módulos desempeñan un papel clave dentro del marco: pueden guiar al modelo hacia una salida adecuada o, en caso contrario, conducirlo a resultados erróneos, lo que resalta su relevancia y necesidad de un cuidadoso diseño.

Como línea de trabajo futuro, se plantea la especificación e implementación de un marco de software en Python que proporcione una estructura completamente reutilizable. Dicho marco deberá facilitar la implementación, evaluación y persistencia de distintas instancias de PAGE, facilitando así su aplicación práctica y la extensión de sus capacidades en nuevos contextos.